\documentclass[journal]{IEEEtran}
\usepackage[square,numbers]{natbib}

\usepackage{multicol}
\usepackage{adjustbox}
\usepackage{comment}
\usepackage[utf8]{inputenc} 
\usepackage[T1]{fontenc}    
\usepackage{hyperref}       
\usepackage{url}            
\usepackage{booktabs}       
\usepackage{amsfonts}       
\usepackage{nicefrac}       
\usepackage{microtype}      
\usepackage{xcolor}         

\usepackage{graphicx}

\title{The Why, When, and How to Use Active Learning in Large-Data-Driven 3D Object Detection for \\ Safe Autonomous Driving: An Empirical Exploration}

\author{Ross~Greer, 
        Bjørk~Antoniussen,
        Mathias~V.~Andersen,
        Andreas~Møgelmose,
        and~Mohan~M.~Trivedi
\thanks{Ross Greer and Mohan Trivedi are with the Laboratory for Intelligent and Safe Automobiles, Department
of Electrical and Computer Engineering, University of California San Diego, La Jolla, CA, 92092 USA e-mail: regreer@ucsd.edu.}
\thanks{B. Antoniussen, M. V. Andersen, and A. Møgelmose are with the Faculty of Electronic Systems, Aalborg University.}}

\begin{document}

\maketitle

\begin{abstract}
Active learning strategies for 3D object detection in autonomous driving datasets may help to address challenges of data imbalance, redundancy, and high-dimensional data. We demonstrate the effectiveness of entropy querying to select informative samples, aiming to reduce annotation costs and improve model performance. We experiment using the BEVFusion model for 3D object detection on the nuScenes dataset, comparing active learning to random sampling and demonstrating that entropy querying outperforms in most cases. The method is particularly effective in reducing the performance gap between majority and minority classes. Class-specific analysis reveals efficient allocation of annotated resources for limited data budgets, emphasizing the importance of selecting diverse and informative data for model training. Our findings suggest that entropy querying is a promising strategy for selecting data that enhances model learning in resource-constrained environments. \\ \\
\indent \textit{Note to Practitioners---}Data-driven systems continue to show state-of-the-art performance across perception tasks in autonomous driving systems. However, the annotation of such data can be prohibitively expensive, with the annotation of non-informative data contributing to wasted labor and computational resources. Active learning systems, like the one we introduce in this paper, allow for the reduction of datasets to the most informative samples, such that the same perception performance can be reached at a limited labeling budget. We relate this data budget to hours of labor to illustrate the importance of this efficiency gain. In this paper, we especially show that these performance effects actually serve a dual purpose, providing enhanced perception of minority classes on limited data. Minority classes stochastically appear most rarely while driving, and safety-critical events occur with such rarity (``long-tail problem") that it is critical to fully utilize such data during model training to improve safety outcomes. 
\end{abstract}

\begin{IEEEkeywords}
active learning, safe autonomous driving, 3D object detection, entropy querying, data-driven perception systems
\end{IEEEkeywords}

\section{Introduction}

Many autonomous driving tasks rely on supervised learning, and task performance under such methods is heavily dependent on accurate, high-volume data annotation. The conventional approach for most autonomous driving tasks, such as 3D object detection \cite{wang2022st, cai2023bevfusion4d, liu2023bevfusion, chen2023focalformer3d, xie2023sparsefusion}, is to ask humans to label (or supervise the labeling of) all data collected in driving, then train learning machines using the labeled data. 

\begin{figure}
    \centering
    \includegraphics[width=.5\textwidth]{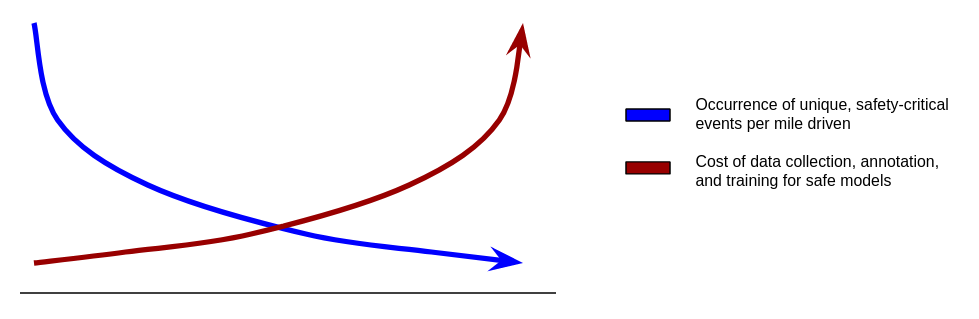}
    \caption{Novel safety-critical events occur with low probability while driving, making data collection of such events an enormous cost, especially since the number of instances required to teach a high-dimensional model scales exponentially with the number of data dimensions. While the left region of this curve may represent scenarios encountered in normal driving, as we progress to the right, we would expect to find not only unexpected driving environments and interactions, but also those of near-miss accidents and catastrophic failures. Collecting real-world data on dangerous accidents (and, at that, sufficient instances of this data to build models via supervised learning in high-dimensional vector spaces) is an extremely challenging task. The blue curve carries the moniker of ``long-tail events".}
    \label{fig:longtail}
\end{figure}

However, such annotation often requires meticulous treatment and expensive labor from expert human annotators \cite{lava}. When the volume of the data grows faster than the available human resources, annotating data becomes a challenging bottleneck to better-performing models. This is especially the case for autonomous driving, where the data itself can be collected quickly and diversely from fleets or even a single vehicle \cite{greer2023champ}. In fact, a German study in autonomous vehicle data estimated the annotation cost to produce direct statistical evidence of reliable AI-perception ranges in the scale of 1.16 trillion to 51,800 trillion Euro -- 14,800 times Germany's gross domestic product! \cite{fingscheidt2022deep, gottschalk2021does} In this research, we explore and evaluate an entropy-based querying active learning solution to this annotation bottleneck with consideration to the multimodal, multitask, and safety-critical nature of intelligent vehicle learning systems. 

\begin{figure*}
    \centering
    \includegraphics[trim=0.3cm 8.2cm 0.3cm 8.2cm, clip, width=\textwidth]{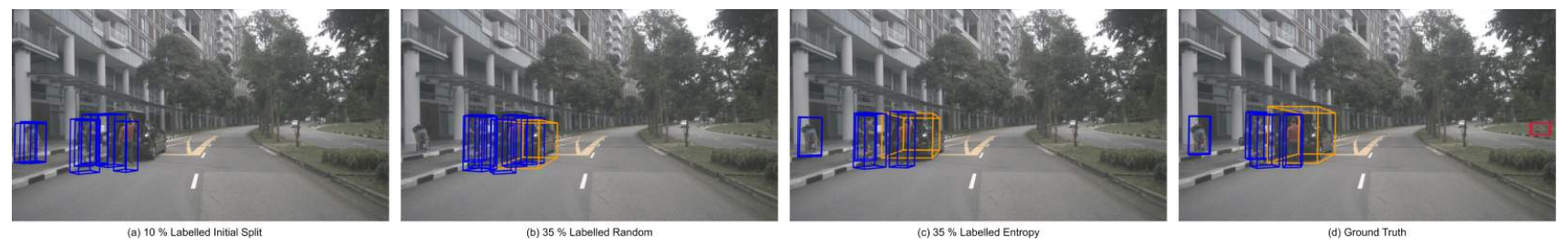}
    \caption{The amount of carefully annotated data available during training is closely tied to the success of the learned model. This is an image from the nuScenes dataset, whose camera and LiDAR measurements are used as input to the BEVFusion 3D Detection model discussed in this paper. When the model is trained with 10\% of the available training data, we can see a high rate of false positive detections throughout the scene, and failure to note even the obvious-but-partially-occluded vehicle. As we increase the training data to 35\% of the available pool, under random sampling, the false positive detections remain confounding, but the pedestrians on the sidewalk are missed altogether, and there is a general difficulty to capture the precise position, size, and orientation of these objects. On the other hand, when using the entropy querying active learning method detailed in this paper, under the same data budget, the pedestrian on the sidewalk is found and the false positive detections are significantly reduced relative to the ground truth. The ground truth, depicted on right, shows the ideal detection, which requires the careful selection of additional data points to further boost trained model performance without incurring expensive demand for extensive data annotation. In this research, we present methods for intelligently querying the available data pool for new training samples using active learning.}
    \label{fig:det_methods}
\end{figure*}

\subsection{Redundancy and Data Imbalance}

As a motivating example, consider a fleet which seeks to gather data in a particular region. By the nature of our roadway system, over time, vehicles will likely encounter the same roads in the same conditions and same context multiple times (e.g. a 5 o'clock rush hour traffic jam on southbound I5 near Exit 26B). For this reason, many data points collected for autonomous driving may be redundant or similar between capture sessions. 

Why is this redundancy, or \textit{data imbalance}, a problem to begin with? When nearly-identical, highly-repeated samples are used to train a model (and distinct samples are significantly less present), the data imbalance can cause the model to overfit parameters to be sensitive to the minor deviations in the over-represented data instead of solving the intended problem -- an issue addressed with active learning \cite{cohn1994improving} \cite{pes2020learning}. Additionally, in a well-designed model trained on a sufficiently diverse dataset, the model learns a latent space which interpolates between encoded samples, allowing the model to generalize to noisy data in the wild \cite{lee2020meta}. While collecting large amounts of data is important, there comes a point when further data collection of similar samples becomes redundant as the learning of the latent space sufficiently covers the real-world pattern for similar samples. This is especially the case when it comes to safety for autonomous driving, as it is not the familiar which poses a risk, but rather encounters with unexpected or novel situations, so-called ``long-tail" (infrequent) driving events. At a practical level, because ML systems optimize over a loss function summed over each training sample, in cases of severe class imbalance, catering to the ``majority" serves to place the learner in a comfortable local minimum of loss. Further, when it comes to safe autonomy, these non-majority cases are often the most significant from a safety standpoint. This challenge is shared with biomedical research, earning the name \textit{curse of rarity}, referring to the difficulty of gathering samples of events that are most likely to cause system safety failures \cite{liu2022curse}. This is also referred to as the ``long-tail problem". 

Data sampling methods are commonly used to overcome data imbalance, such as random under-sampling (to remove majority cases from training data), and random over-sampling (having under-represented classes appear more frequently during training). In principle, standard data augmentation serves this same purpose, but on the basis that the collected data under-represents the variance of the complete population of data. Naturally, augmentation methods can be applied to minority-class data to build a stronger representation within a training dataset. However, here we seek solutions which add more to a model's knowledge than crafted re-use of existing training data, such that a system can continually learn from new examples, finding ``useful novelty" through examining the entropy of  considered data \cite{dubnov2023deep}. 

\subsection{Dealing with High-Dimensional Data}

In addition to data imbalance, data for intelligent vehicle tasks tends to be high-dimensional. For example, a typical testbed may be collecting data along dimensions of time, arrays of pixels from 2D spatial cameras, sweeps of 3D spatial lidar measurements, and a variety of additional sensors such as GPS, INS, and CAN.  

By learning an expansive low-to-high-level feature set, this scale and variety of information has proven to be helpful towards a variety of tasks such as lane detection \cite{abualsaud2021laneaf}, vehicle and VRU detection and tracking \cite{9304793}, traffic sign and light classification \cite{salient, greer2023salient, greer2023robust}, trajectory prediction \cite{lefevre2015learning, greer2021trajectory}, vehicle landmark identification \cite{GREER2024}, driving maneuver and driver style classification \cite{doshi2010examining}; such tasks are important not only towards autonomous driving, but also towards effectiveness of ADAS systems \cite{balachandran2015predictive}. While this data provides a wealth of information to learn from, the infamous ``curse of dimensionality" puts systems at risk of improperly fitting models to complex data (requiring exponential amount of increased data with each new dimension introduced). Further, even annotating this data at a high-quality, frame-by-frame, pixel-by-pixel, voxel-by-voxel level is a monumental task, near impossible to complete exhaustively given resource constraints and costs in human annotation, discussed further in later sections. 

\begin{table}
\caption{Percentage of nuScenes 3D object dataset possible to be annotated by 40 hours of work, calculated from rate estimates in recent research.}
\label{nuscenes}
\begin{adjustbox}{width=0.49\textwidth}
\begin{tabular}{c|c}
    Method & \% of nuScenes objects annotated in 40 Hours (est.) \\
    \hline \hline 
    \rule{0pt}{2ex} 3D-BAT \cite{zimmer20193d} & 6.86\% \\
    \hline
    \rule{0pt}{2ex} Lee et al. \cite{lee2018leveraging} & 2.78\% \\
    \hline
    \rule{0pt}{2ex} Without assistance average \cite{liu2022map} & 0.09\% \\
    \hline
    \rule{0pt}{2ex} With assistance average \cite{liu2022map} & 0.34\% \\
    \hline
\end{tabular}
\end{adjustbox}
\end{table}

\begin{table}
\caption{Percentage of nuScenes 3D object dataset annotated by 40 hours of work per week, with the number of weeks shown in the left column, calculated using the average rate of Table \ref{nuscenes}. We use this dataset size as the allowed size of our training sets to evaluate the effectiveness of active learning approaches.}
\label{nuscenesprogress}
\begin{center}
\begin{adjustbox}{width=0.40\textwidth}
\begin{tabular}{c|c}
    Weeks of Annotation & \% of nuScenes dataset in Training Pool \\
    \hline \hline 
    \rule{0pt}{2ex} 1 & 2.52\% \\
    \hline
    \rule{0pt}{2ex} 2  & 5.04\% \\
    \hline
    \rule{0pt}{2ex} 3 & 7.56\% \\
    \hline
    \rule{0pt}{2ex} 4 & 10.08\% \\
    \hline
    \rule{0pt}{2ex} 5 & 12.60\% \\
    \hline
    \rule{0pt}{2ex} 6  & 15.12\% \\
    \hline
    \rule{0pt}{2ex} 7 & 17.64\% \\
    \hline
    \rule{0pt}{2ex} 8 & 20.16\% \\
    \hline
    \rule{0pt}{2ex} 9 & 22.68\% \\
    \hline
    \rule{0pt}{2ex} 10 & 25.20\% \\
    \hline
\end{tabular}
\end{adjustbox}
\end{center}
\end{table}

In essence, much of machine learning involves reducing the dimensionality of data from its high-dimensional observed form to a task-useful form. Sometimes we do this before the data enters the learning mechanism (e.g. pre-processing the data by selecting features to learn from), sometimes we do this inside the learning mechanism (e.g. an early bottleneck layer in a neural network, which learns lower-dimensional encodings of feature combinations). Sometimes we do this explicitly (e.g. extract particular features, such as one color channel for a task like brake light extraction \cite{GREER2024}), often termed \textit{selecting}, other times letting the system learn the features (e.g. neural network which outputs a low-dimensional vector for system inference \cite{greer2023multi}), often termed \textit{mapping}.

\begin{table*}
    \caption{Comparison of previous research in Active Learning for 3D Object Detection in Autonomous Driving Datasets}
    \centering
    \begin{tabular}{p{6cm}|p{1.4cm}|p{1.1cm}|p{7.5cm}}
        Active Learning Methods & Datasets & Modalities & Insights \\
        \hline \hline 
        Entropy, Monte Carlo dropout, ensemble learning  & KITTI & Camera, LiDAR & Can save up to 60\% of labeling efforts for same performance \cite{feng2019deep} \\ \hline 
        Class Entropy and Spatial Uncertainty & Private & LiDAR & Importance of both classification and spatial uncertainty \cite{moses2022localization} \\ \hline 
        Kernel coding rate & KITTI, Waymo & LiDAR & 44\% box-level annotation costs savings without compromising performance \cite{luo2023kecor} \\ \hline 
        Sensor consistency-based selection score, LiDAR guidance as semi-supervision for monocular detection & KITTI, Waymo & Camera & 17\% savings in labeling costs, top performance in BEV monocular object detection official benchmarks with 2.02 AP gain \cite{hekimoglu2023multi, hekimoglu2024monocular} \\ \hline 
        3D consistency of bounding box predictions in both semi-supervised and active learning & KITTI & LiDAR & Improves from baseline by more than 60\% with only 1500 annotated frames \cite{hwang2023joint} \\ \hline 
        Consensus score variation ratio, sequential region-of-interest matching & KITTI & Camera, LiDAR & Saves 35\% of labeling efforts \cite{schmidt2020advanced} \\ \hline 
        Bi-domain active learning, diversity-based sampling & KITTI & LiDAR & Gains on cross-domain settings; retraining Waymo-trained model on just 5\% of KITTI data outperforms 100\% KITTI-trained model \cite{yuan2023bi3d} \\ \hline 
        Uncertainty sampling & Astyx & Radar, Camera, LiDAR & Semi-automatic labeling for efficient dataset creation  \cite{meyer2019automotive} \\ \hline 
        Augmentation, dropout, insertion, deletion & KITTI & LiDAR & Practical method for fast annotation \cite{meng2021towards} \\ \hline  
        Semi-supervised co-training on prediction disagreement & KITTI, Waymo & Camera & Semi-supervised co-training clearly boosts detection accuracy in regimes where the training size is just 5-10\% of the pool \cite{villalonga2020co} \\ \hline 
        Ego-pose distance-based sampling & Navya3DSeg & LiDAR & Heuristic-free method; outperform random sampling \cite{almin2023navya3dseg} \\ \hline 
        Bayesian surprise (KL divergence) & AGV Anomaly Dataset & LiDAR & Effective in warehouse environment anomaly detection; may be applied as AL to identify novel data \cite{ccatal2020anomaly} \\ \hline 
        Uncertainty sampling & Private & LiDAR, Camera & Effective for identifying road damage \cite{chen2022gocomfort} \\ \hline 
        Spatial and temporal diversity-based sampling & NuScenes & LiDAR & Annotation costs vary between scenes; diversity methods are effective and allow warm start \cite{liang2022exploring} \\ \hline
        \textbf{Classification Entropy Querying} & \textbf{NuScenes} & \textbf{LiDAR, Camera} & \textbf{Outperforms random sampling, reduces intra-class performance difference, learning of minority classes (this research)} \\ \hline
    \end{tabular}
    \label{tab:al_det}
\end{table*}

In addition to implications toward the theoretical limits of a systems ability to learn, high-dimensional data also contributes to a lack of explainability in systems, and complicates the process of safety regulation on a practical level. Techniques in intelligent data selection and feature extraction help to resolve these challenges, but as information is discarded, a tradeoff is induced between system performance and system explainability. Pes et al. \cite{pes2020learning} categorize three types of feature selection methods: 
\begin{itemize}
    \item Filter methods, which remove data according to some non-learned criteria,
    \item Wrapper methods, which essentially search over different feature subsets to optimize performance, and
    \item Embedded methods, which, critically, make use of learning algorithm internal information in the process of searching for optimal features. For example, while a wrapper method might make use of system accuracy over a test set to select a best feature set, an embedded method may examine the uncertainty values of logits during classification to drive its selection criteria. 
\end{itemize}

As expected, filter methods bear the least computational cost, but show the most constrained performance (albeit, sometimes this constrained performance may be sufficient towards a task). In this research, we explore an embedded method, accepting increased computational complexity to enhance model performance.



\subsection{Using Active Learning}

Active learning is the process by which a learning system interactively selects which data points should be added from the unlabeled data pool to the labeled training set, assisted by the intervention of a human expert providing associated annotations. Within this framework, in the case of classification tasks, we consider the \textit{information gain} of a new datum to be a measure of the decrease in entropy when that datum is added to the training set.

This problem is therefore twofold: (1) for model cost and performance, a large set of these non-informative data points increases the time and decreases effectiveness of the training process and model tuning, and (2) for annotation cost, in situations where a data corpus has high levels of redundancies, annotating all collected data may waste a lot of human resources on non-informative samples.

\section{Related Research}



Cohn et al. engage in a particular style of active learning as concept learning via queries, by which the learner requests from an oracle a label for a particular sample \cite{cohn1994improving}. In particular, their work examines the effectiveness of such methods in improving generalization behavior. One of the goals in active learning is to label a small subset of collected unlabeled data so as to maintain or achieve better performance given the cost of labeling or requesting human oracle.
Conventional query strategies usually evaluate informativeness based on handcrafted functions or heuristic selection methods, such as query-by-committee \cite{seung1992query}, uncertainty sampling \cite{lewis1994heterogeneous,scheffer2001active}, region of uncertainty and version space \cite{cohn1994improving}, and expected model change \cite{settles2007multiple}. Empirical studies \cite{schein2007active,settles2008analysis} have shown that the best strategy or informativeness measure may be application specific. Moreover, the effectiveness of such heuristic methods is limited and varies across different datasets. 

Due to the variability in datasets, models, and query selection methods, it is difficult to form a noticeable consensus for the state of the art in active learning. Accordingly, through this paper, we show the clear utility of one such method towards the detection safety goals of autonomous driving systems. Early works in the literature applying active learning in autonomous driving tasks mostly utilized handcrafted features such as Haar wavelets and the histogram of oriented gradients on SVMs or Adaboost \cite{sivaraman2010general,sivaraman2014active,satzoda2015multipart}. As deep learning became a dominant approach in computer vision \cite{krizhevsky2012imagenet}, more works have resorted to DNNs as models in active learning to further boost performance. In \cite{singh2020deep}, four active learning methods (sum of entropy, maximum entropy, average entropy, and Monte Carlo dropout) are applied to the Apollo Synthetic dataset and Waymo Open dataset on 2D object detection and instance segmentation tasks, using R-CNNs appropriate for each task, and finding that active learners beat baselines in these autonomous driving tasks, and that summation-entropy learners tend to bring forward samples with the most instances, which seem to have the strongest effect on learning. While these insights are valuable, in this research, we focus on the task of 3D object detection, reflecting the need for vehicles to recognize an object's relative position for purposes of safe planning; accordingly, our discussion of related works will continue with active learning towards this task. We highlight relevant literature towards effective detection and efficient annotation of such datasets in Table \ref{tab:al_det}, and discuss particular methods in the following paragraphs. 

In \cite{feng2019deep}, Feng et al. use active learning to find the fewest number of labeled training samples to improve the performance of 3D object detection by convolutional neural networks (CNNs) trained on LiDAR point clouds, using Monte Carlo Dropout and Deep Ensembles to measure entropy in predictive labels and mutual information between model weights and class predictions. Moses et al. \cite{moses2022localization} coin a ``\textit{LOCAL}" acquisition function, utilizing both classification and localization-based uncertainty and summing values across all objects in a sample as inclusion criteria. They adapt the exclusive Basic Sequential Algorithmic Scheme (BSAS) clustering scheme for per-object matching to allow for entropy calculation, and use variance of spatial estimation as measure of spatial uncertainty. However, their training and evaluation is carried out on a limited 41 LiDAR point clouds of data from a private, government-owned airborne-collected dataset, and they point out the important difference in scale compared to autonomous driving datasets such as KITTI \cite{geiger2012we}, Waymo, and nuScenes.  
Luo et al. \cite{luo2023kecor} show that maximizing the kernel coding rate as criteria for data selection can strongly outperform most generic (task-agnostic) active learning methods, and marginally improves over task-specific active learning methods for 3D detection, at lower running time than near performers.
Hekimoglu et al. \cite{hekimoglu2022efficient} use active learning on a monocular-input for the 3D detection task, quantifying uncertainty using (1) the variance of predicted Gaussian localizations, and (2) the variance in predicted position when an image undergoes a variety of intensity and sharpness transforms to form a query-by-committee, and perform experiments using a fixed training size, showing that the combinations of data augmentation query-by-committee and heatmap uncertainty lead to clear improvement over random sampling. Hekimoglu et al. are later the first to use a teacher-student paradigm for active learning data selection and semi-supervised training, this time combining LiDAR measurement with monocular images to form this teacher-student relation, and setting a new state of the art for ``monocular" (since the LiDAR is technically used without label) 3D object detection on KITTI \cite{hekimoglu2024monocular}. 
Hwang et al. \cite{hwang2023joint} exploit the ability to localize 3D objects under flips, rotations, and scalings so that unlabeled data can be used to train the model to be consistent in assessing object locations, using this value as both an additional training term and uncertainty measurement towards active learning. These papers are all united on the theme that active learning leads to higher 3D detection performance at lower data budgets, shown in a general sense on a limited number of object classes. 

From our search, Liang and et al. \cite{liang2022exploring} provide the only prior investigation of active learning on the nuScenes dataset. While we study uncertainty-based active learning in this research, Liang et al. study diversity-based active learning, finding that spatial and temporal diversity of samples are effective strategies. They importantly highlight the differences of annotation costs being variable between scenes, due to the varying number of objects that may appear in each; accordingly, they define the annotation budget by a combined scene-object formulation. They also hypothesize that these entropy-based methods may introduce redundant samples in a scene, since having a high-entropy class at any one pooling round would likely identify all members of that class to be high entropy, when a smaller representative amount would suffice for learning. Further, their diversity-based active learning approach allows for a ``warm start" to their base training pool, as the diversity criteria can be established without a trained model. Under their annotation budget, the entropy-based method appears to underperform compared to random sampling (and, this makes sense given that a scene's entropy is formed by the sum of the detected object entropies). However, we do recommend that entire scenes be annotated at once (even if highly crowded), due to the difficult task of the model to identify all objects within any annotated scene; state-of-the-art models are not trained to look for single objects in a field of many, but rather to identify all instances simultaneously, and the task of identifying instances within a crowd warrants appropriate data. Accordingly, we show that at the scene-sampling level budget, entropy-driven active learning actually does exceed a random baseline.

We point out key differences between our research and the research of \cite{feng2019deep}, \cite{luo2023kecor}, \cite{hekimoglu2022efficient}, \cite{hekimoglu2024monocular}, \cite{hwang2023joint}, and \cite{liang2022exploring}:
\begin{itemize}
    \item We experiment over the nuScenes  \cite{NuScenesPaper}, while other works experiment on KITTI \cite{feng2019deep, luo2023kecor, hekimoglu2022efficient, hekimoglu2024monocular, hwang2023joint} and Waymo \cite{luo2023kecor, hekimoglu2024monocular}; by experimenting on an additional strongly-established dataset, we further enhance their case for the benefits of entropy-driven querying and active learning in autonomous driving. 
    \item Accordingly, while the KITTI and Waymo-based approaches \cite{feng2019deep, luo2023kecor, hekimoglu2022efficient, hekimoglu2024monocular, hwang2023joint} divide objects into five or less classes (for example, small vehicle, human, truck, tram, and miscellaneous), we divide objects into 10 classes\footnote{Pedestrian, Bicycle, Car, Bus, Construction Vehicle, Motorcycle, Barrier, Traffic Cone}, better capturing the distribution of minority classes and the effects of active learning on less-represented data.
    \item Some of the above prior works do not include orientation \cite{feng2019deep, liang2022exploring} or classification \cite{hekimoglu2022efficient} of objects in their detection. These attributes are important for the purposes of understanding possible direction-of-travel and behavioral patterns for an object \cite{mogelmose2015trajectory}. We include and evaluate these predictions in our network output. 
    \item \cite{feng2019deep} uses ground-truth and pre-trained image 2D detectors in their 3D detection pipeline, while \cite{luo2023kecor, hwang2023joint, liang2022exploring} utilize LiDAR only and \cite{hekimoglu2022efficient} utilizes monocular camera only. By contrast, we \textit{train} our image-based 2D detector as part of a two-stage (image + LiDAR) network; thus, active learning decisions influence the complete network performance. 
\end{itemize}


We create an active learning framework for autonomous driving to jointly minimize redundant, expensive annotation while avoiding the risk introduced by domain adaptations and overfitting. Such an approach allows autonomous vehicles to efficiently learn new knowledge for unseen environments under constrained resources.

\subsection{How long does it take to annotate 3D bounding boxes?}

3D object detection is a very relevant and important task to autonomous driving because unlike 2D object detection, the object's position and orientation in space is inferred. However, the task of drawing 3D bounding boxes to train models for such tasks can be more time consuming than 2D annotation. In this section, we highlight just how expensive this data can be to make a case for active learning as a cost-reducing measure so these systems can be developed safely at scale. 

To assist in this annotation task, tools such as Zimmer et al.'s 3D-BAT \cite{zimmer20193d} have been developed for semi-automatic labelling. In the 3D-BAT test case, they find that the most efficient expert human annotator is able to use the system to annotate approximately 57 objects per minute, and the average among users is approximately 40 objects per minute. However, IoU with ground truth is very low for these fast annotations, with the best annotator reaching only around 20\%. Lee et al. design a system where annotators provide object anchor clicks to generate instance segmentation results in 3D, reporting 3.7 seconds per bounding box \cite{lee2018leveraging}. To motivate their auto-labelling system MAP-Gen, Liu et al. report statistics that an experienced annotator takes around 114 seconds per 3D bounding box, and those using a 3D object detector assistant around 30 seconds \cite{liu2022map}. While auto-labelling may eventually be a viable solution toward massive data annotation, here we emphasize the importance of expert annotators in the loop for the purpose of human-validated safety in such a risky domain. 

NuScenes contains 1.4M camera images and 390k LIDAR sweeps of driving data, originally labeled by expert annotators from an annotation partner. 1.4M objects are labelled with a 3D bounding box, semantic category (among 23 classes), and additional attributes. In Table \ref{nuscenes}, we form estimates of the portion of nuScenes dataset that annotators utilizing above-described methods could annotate in 40 hours, again noting that the quality of annotation for some of these methods is substandard. 

\begin{figure*}
    \centering
    \includegraphics[width=.9\textwidth]{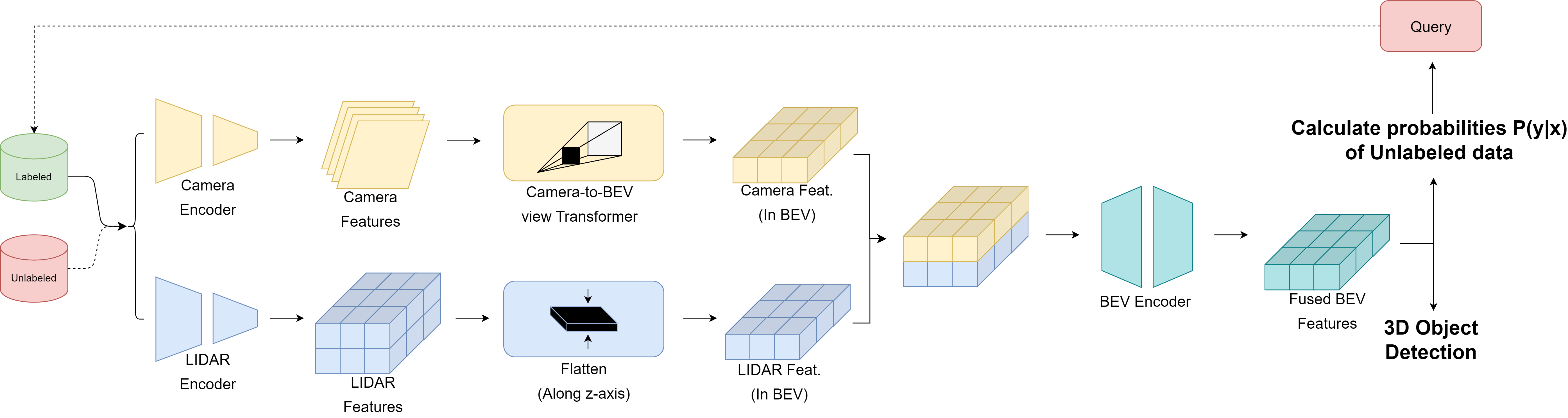}
    \caption{
    An illustration of Active Learning setup with the BEVFusion model. 
    }
    \label{fig:AL_BEVFusion}
\end{figure*}

\begin{table}[]
    \centering
    \caption{Class Frequencies in NuScenes, ordered most to least present.}
    \begin{tabular}{c|c}
        Class & Frequency (\%) \\
        \hline 
        Car & 42.30 \\ 
        Pedestrian & 19.05 \\
        Barrier & 13.04 \\
        Traffic Cone & 8.40 \\
        Truck & 7.59 \\
        Trailer & 2.13 \\
        Bus & 1.4 \\
        Construction Vehicle & 1.26 \\ 
        Motorcycle & 1.08 \\
        Bicycle & 1.02 \\
    \end{tabular}
    \label{tab:freq}
\end{table}

Though this paper demonstrates the utility of active learning towards the task of 3D object detection, we would like to stress that this paper is not about improved 3D object detection, but rather about systematically selecting data in a way that improves model learning under limited resources. There are many additional tasks in autonomous driving beyond 3D object detection; for example, Motional has accompanying semantic visual and LiDAR segmentation tasks, which are even more time-intensive during annotation (for example, Schmidt estimates up to 90 minutes to fully segment an autonomous vehicle domain image \cite{schmidt2019crowdproduktion}). The benefits demonstrated on our sample task are applicable towards other tasks; active learning is used to increase efficient utility of data towards improving any task model, especially in the cases of multi-task active learning frameworks \cite{hekimoglu2023multi, finn2018probabilistic}.

\section{Data Methods}
Because the rate of newly collected data is faster than the rate of annotation, prioritizing data for learning new knowledge is expected to boost performance in a more optimal rate per datum. Therefore, we formulate the autonomous driving tasks as pool-based active learning problems \cite{lewis1994sequential}. We assume that large collections of unlabeled data are collected continuously in the pool and associate queries for the accurate annotation by expert human annotators with some costs. To minimize the total cost while maximizing the autonomous driving performance, our proposed algorithms only request humans to annotate data points when they are novel to the existing dataset and influential to the current model. The other data points are assigned with the label generated by the current model or have their annotation delayed. For evaluation, the model is trained with a few steps in each cycle based on the union of the requested labels and a subset of assigned labels of data points.

\subsection{Active Learning for NuScenes}

NuScenes comprises 1000 scenes. In order to maintain complete control over the scenes within the dataset, we will be making slight adjustments to the fundamental database setup. These modifications are necessary to accommodate the presence of unlabeled data and the computations associated with active learning queries. The specific adjustments will depend on the selected method. This alteration is a crucial step in the process of sampling underrepresented data from the current labeled pool.

Towards reproducability of our methods, throughout the training and testing of the chosen model we will use the \textit{trainval} split of the dataset, which containes 850 scenes. We will split this into labeled, unlabeled and validation subsets, where the validation set will contain 150 scenes used to evaluate and test the model. We will discard the provided \textit{test} subset for our experiments, as the labels are not provided by the creators. 

The remainder of the scenes in \textit{trainval} will initially be part of the unlabeled subset and iteratively be sampled approximately 5\% at the time into the labeled set. This process will proceed until models have been trained on the labeled subset containing up to 50\% of the original scenes present in the \textit{trainval} dataset.



\subsection{Baseline: Random Sampling}

We create a baseline budget using the average of the statistics surveyed in Table \ref{nuscenes}, or 2.52\% of the nuScenes dataset annoted with a 40-person-hour labelling budget. We create 10 iterative batches of such labels, representing in a figurative sense the amount that one (very dedicated) annotator might label over 10 weeks, shown in Table \ref{nuscenesprogress}.

For each baseline trial, we randomly sample a percentage of scenes described in Table \ref{nuscenesprogress} and train the model to N epochs. We will start with 10.08\% scenes and add 5.04\% for every round representing a start with 4 weeks worth of work and an increase of 2 weeks worth of work for every sampling round.   

\subsection{Active Learning Method: Entropy Querying}

We aim to investigate the implications of utilizing a commonly employed uncertainty measure for sampling from an unlabeled data pool \cite{feng2019deep}, \cite{hekimoglu2022efficient}, \cite{hekimoglu2024monocular}, \cite{hwang2023joint}.

While certain methods, like "least confidence" and "smallest margin," derive their acquisition function based on individual or paired confidence values across all semantic classes, our specific focus lies on the "entropy querying" method. This method takes into account a model's uncertainty across all conceivable classes. Our objective is to uncover potential enhancements that the entropy query method could bring about, given that the informativeness measure is determined by comparing a sample's probability of belonging to a class across all possible classes. \cite{uncertaintyALL}



This process starts by conducting inference on the unlabeled subset and strategically selecting samples found to be the most informative. The criterion for informativeness is determined by the entropy scores associated with each sample. These scores are calculated, generally, using the formula expressed in Equation \ref{entropy_equ}. 

\begin{equation}
    \Phi_x = \sum_y P(y|x)\log_2P(y|x)
    \label{entropy_equ}
\end{equation}

In the equation, $\Phi_x$ represents the entropy score for a given sample $x$. The calculation involves the summation over all possible class labels $y$, where $P(y|x)$ represents the probability of class $y$ given the input $x$. The resulting entropy score serves as a quantitative measure of uncertainty, guiding the selection of samples for active learning.

By adopting the entropy sampling approach, we aim to enhance our understanding of its impact on the selection process within the context of 3D datasets. The utilization of entropy scores provides a nuanced perspective on uncertainty, enabling the selection of samples that contribute most significantly to the model's learning process.

\begin{table*}
\caption{Performance across standard 3D object detection metrics at different training dataset sizes, training by Random Sampling and Entropy Querying.}
\label{nuscenesprogress_random}
\begin{adjustbox}{width=\textwidth}
\begin{tabular}{c|c|c|c|c|c|c|c|c|c|c|c|c|c|c|c}
    Round & Pool & \multicolumn{2}{c|}{mAP $\uparrow$} & \multicolumn{2}{c|}{mATE $\downarrow$} & \multicolumn{2}{c|}{mASE $\downarrow$} & \multicolumn{2}{c|}{mAOE $\downarrow$} & \multicolumn{2}{c|}{mAVE $\downarrow$} & \multicolumn{2}{c|}{mAAE $\downarrow$} & \multicolumn{2}{c}{NDS $\uparrow$} \\
    \hline \hline 
        
    \hline
    \multicolumn{2}{c|}{} & Random & Entropy & Random & Entropy & Random & Entropy & Random & Entropy & Random & Entropy & Random & Entropy & Random & Entropy \\
    \hline
    \hline
    \rule{0pt}{2ex} 1  & 10\% & 0.3095 & \textbf{0.3106} & 0.4665 &\textbf{0.4588} & \textbf{0.3494} & 0.3669 & 1.108 & \textbf{1.030} & \textbf{1.236} & 1.420 & 0.3794 & \textbf{0.3187} & 0.3353 & \textbf{0.3409} \\
    \hline
    \rule{0pt}{2ex} 2 & 15\% & 0.3419 & \textbf{0.3639} & 0.4392 & \textbf{0.4144} & 0.3397 & \textbf{0.3386} & 0.9418 & \textbf{0.8909} & \textbf{1.223} & 1.347 & 0.3095 & \textbf{0.3074} & \textbf{0.3679} & 0.3868 \\
    \hline
    \rule{0pt}{2ex} 3 & 20\% & 0.380 & \textbf{0.4041} & 0.4041 & \textbf{0.3994} & 0.3503 & \textbf{0.3270} & 0.8296 & \textbf{0.8131} & 1.317 & \textbf{1.060} & 0.3017 & \textbf{0.2955} & 0.4014 & \textbf{0.4185} \\
    \hline
    \rule{0pt}{2ex} 4 & 25\% & \textbf{0.4236} & 0.4217 & 0.3921 & \textbf{0.3786} & \textbf{0.3136} & 0.3319 & 0.7685 & \textbf{0.6780} & \textbf{0.8695} & 0.9803 & \textbf{0.277} & 0.2942 & \textbf{0.4497} & 0.4446 \\
    \hline
    \rule{0pt}{2ex} 5 & 30\% & 0.4494 & \textbf{0.4557} & 0.3713 & \textbf{0.3552} & \textbf{0.3112} & 0.3169 & 0.6989 & \textbf{0.6563} & 0.7764 & \textbf{0.7106} & 0.2485 & \textbf{0.2287} & 0.4841 & \textbf{0.5011} \\
    \hline
    \rule{0pt}{2ex} 6 & 35\% & 0.4474 & \textbf{0.4676} & \textbf{0.3498} & 0.3679 & 0.3168 & \textbf{0.3066} & 0.6569 & \textbf{0.6152} & 0.8830 & \textbf{0.6354} & 0.2941 & \textbf{0.2324} & 0.4736 & \textbf{0.5181} \\
    \hline
    \hline 
    \rule{0pt}{2ex} SOA & 100.00\% & \multicolumn{2}{c|}{0.750} & \multicolumn{2}{c|}{-} & \multicolumn{2}{c|}{-} & \multicolumn{2}{c|}{-} & \multicolumn{2}{c|}{-} & \multicolumn{2}{c|}{-} & \multicolumn{2}{c}{0.761} \\
    \hline
\end{tabular}
\end{adjustbox}
\end{table*}

\subsection{BEVFusion Model for 3D Object Detection}
For the purpose of designing and experimenting on data selection and learning schemes, in this paper we consider the fundamental driving task of 3D object detection. This is an essential task for obstacle avoidance and path planning. 

More specifically, we consider the recent BEVFusion approach to 3D object detection \cite{bevFusion}. At the time of writing, this method holds third place in the NuScenes tracking challenge and seventh place in the detection challenge, with newer variants of the BEVFusion architecture populating additional high rankings. While there are many techniques to find a unified representation of image and LiDAR data, LiDAR-to-Camera projection methods introduce geometric distortions, and Camera-to-LiDAR projections struggle with semantic-orientation tasks. BEVFusion is meant to create a unified representation which maintains both geometric structure and semantic density. 

The Swin-Transformer \cite{swinT} is used as the image backbone, while VoxelNet \cite{VoxelNet} is used as the LIDAR backbone. To create the bird's-eye-view (BEV) features for images, first a Feature Pyramid Network (FPN) \cite{FPN} is applied to fuse the multi-scale camera features. This produces a feature map 1/8 of the original size. After this, images are downsampled to 256x704 and the LiDAR point clouds are voxelized to 0.075m to get the BEV features needed for object detection. These two modalities are fused using a convolution-based BEV encoder to prevent local misalignment between LiDAR-BEV features and camera-BEV features under depth estimation uncertainty from the camera mode. The full architecture with active learning can be seen in \ref{fig:AL_BEVFusion}.




\subsection{Explanation of nuScenes Metrics}

We summarize here some common metrics in 3D object detection for conceptual description, and direct the reader to the nuScenes documentation for implementation thresholds and class-specific details:

\begin{itemize}
    \item Mean Average Precision (mAP): for the nuScenes dataset, AP is computed by taking the 2D center distance on the ground plane, filtering predictions beyond a certain threshold, and integrating  the recall-precision curves for values over 0.1. These values are averaged over match thresholds of {0.5, 1, 2, 4} meters, and then averaged across classes. 
    \item Average Translation Error (ATE): Euclidean center distance in 2D in meters.
    \item Average Scale Error (ASE): $1 - IOU$ after aligning centers and orientation.
    \item Average Orientation Error (AOE): Smallest yaw angle difference between prediction and ground-truth in radians. 
    \item Average Velocity Error (AVE): Absolute velocity error in m/s.
    \item Average Attribute Error (AAE): Calculated as $1 - acc$, where $acc$ is the attribute classification accuracy.
\end{itemize}

These metrics are all positive (or zero) valued, and translation and velocity errors can grow unbounded. For metrics presented in this paper, we take a mean over all classes when presenting general statistics in Table \ref{nuscenesprogress_random}, and also examine per-class performance to observe the effects of active learning on minority classes in further analysis.

\section{Experimental Evaluation}
Experiments are conducted to test if entropy sampling performs better than random sampling. The initial dataset contains approximately 10\% of the original dataset, we add approximately 5\% of data for each subsequent round of training. 

A single round involves training one model with six epochs on the current labeled training set. Following this training phase, the checkpoint file from the last round is employed to perform inference on the unlabeled dataset pool. Thereafter, the employed active learning method will be used on the obtained results. This process identifies the samples to be included in the labeled training dataset for the subsequent round. Each experiment will involve six rounds, as seen in Table \ref{nuscenesprogress_random}. We note that in general, the more training data sampled, the stronger the model learns to generalize to real-world test data.

\begin{figure*}
    \centering
    \includegraphics[trim=0.3cm 8.2cm 0.3cm 8.2cm, clip, width=.95\textwidth]{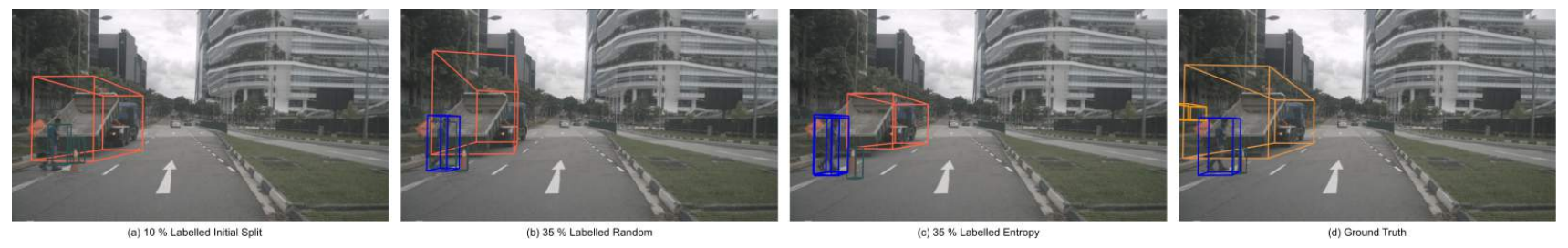}
    \includegraphics[trim=0.3cm 8.2cm 0.3cm 8.2cm, clip, width=.95\textwidth]{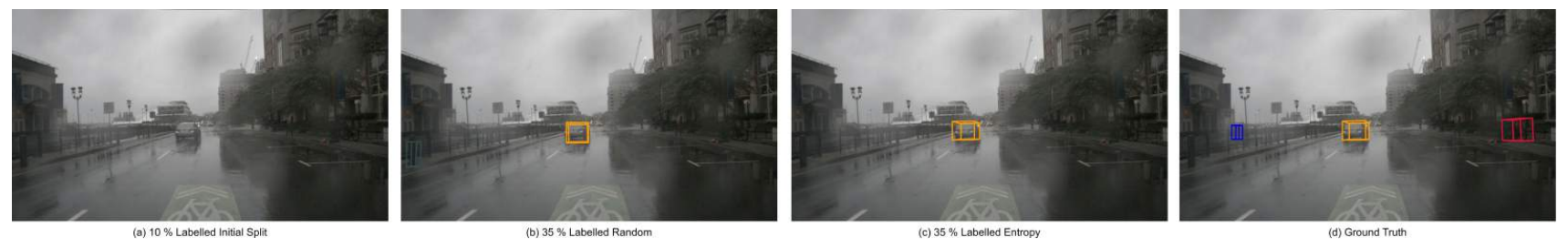}
    \includegraphics[trim=0.3cm 8.2cm 0.3cm 8.2cm, clip, width=.95\textwidth]{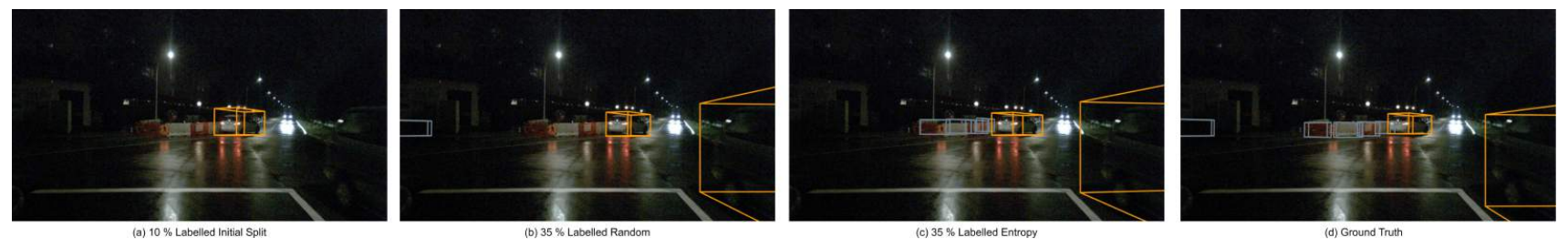}
    \includegraphics[trim=0.3cm 8.2cm 0.3cm 8.2cm, clip, width=.95\textwidth]{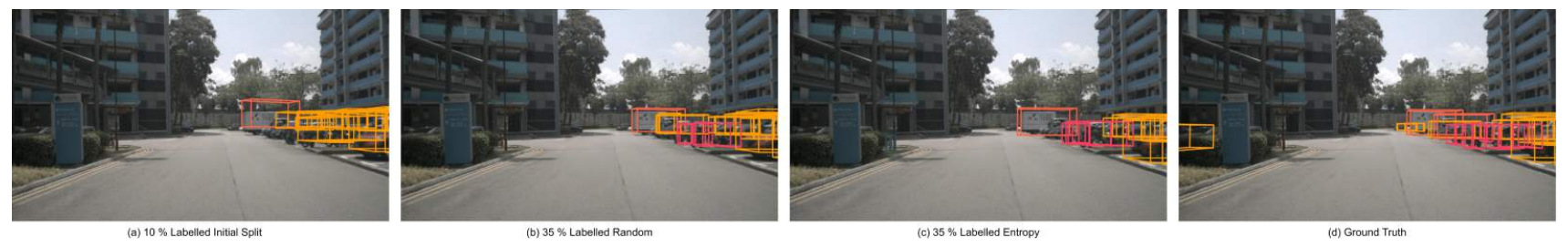}  
    \includegraphics[trim=0.3cm 8.2cm 0.3cm 8.2cm, clip, width=.95\textwidth]{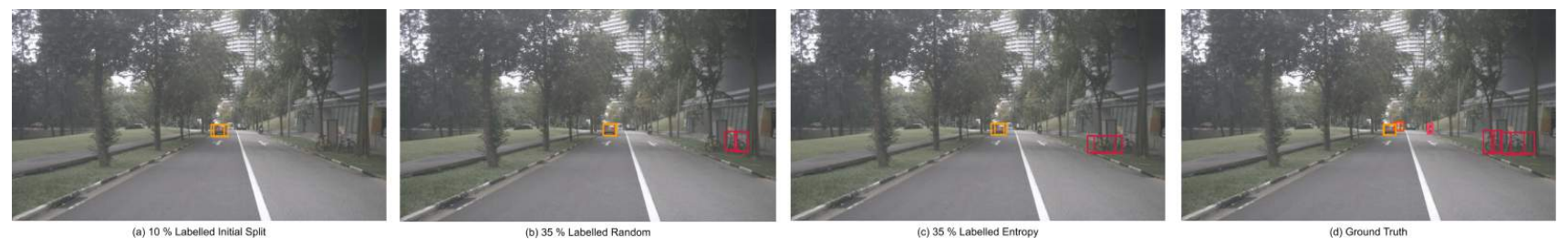}
    \includegraphics[trim=0.3cm 8.2cm 0.3cm 8.2cm, clip, width=.95\textwidth]{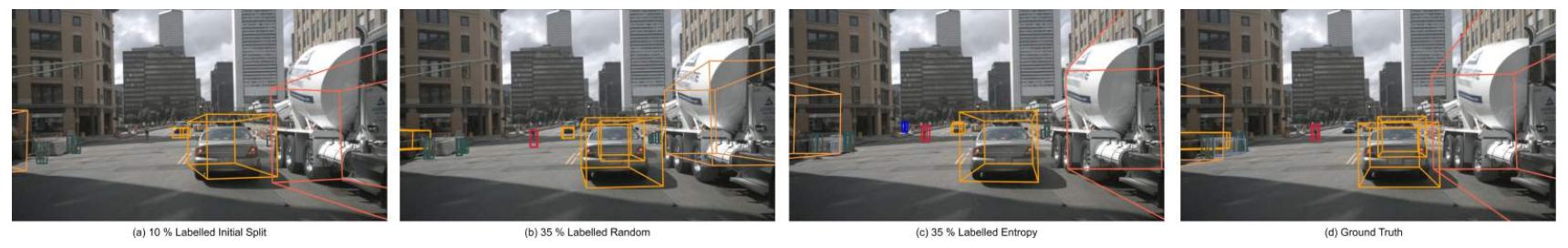}
    \includegraphics[trim=0.3cm 8.2cm 0.3cm 8.2cm, clip, width=.95\textwidth]{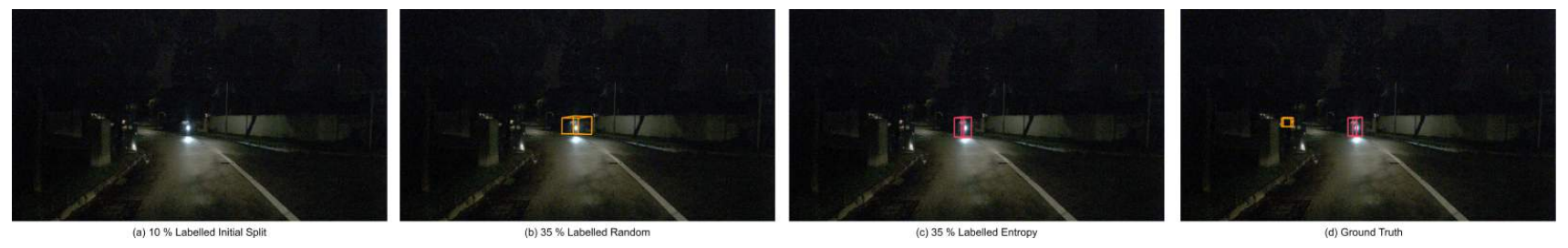}
    \caption{Qualitative examples from nuScenes, comparing the initial training on a randomly selected 10\% of nuScenes, followed by random sampling to 35\% of nuScenes versus 35\% selected using entropy queries, and finally the ground truth annotations. Different color boxes refer to different object classes. A few notable observations under entropy-driven AL: in the first and second row, we see a better handling of orientation-error; in the third row (night), the barrier class is more readily detected; in the fifth row, the presence shape of the bicycles are better inferred; in the sixth row, the truck class and size is correctly inferred; finally, in the last row, the nearby bicycle is detected and correctly classified, where it is missed altogether or mistaken to be a car via random sampling.}
    \label{fig:qual2}
\end{figure*}

The Active Learning strategy dominates on 26 of the 35 checkpoints and metrics in Table \ref{nuscenesprogress_random}. A sampling of qualitative examples are provided in Figure \ref{fig:qual2}. 

\begin{table*}
    \centering
    \caption{Merged table with samples from random and entropy sampling.}
    \begin{adjustbox}{width=\textwidth}
    \begin{tabular}{c|c|c|c|c|c|c|c|c|c|c|c|c}
    \hline
        Data [\%] & \multicolumn{2}{c|}{10} & \multicolumn{2}{c|}{15} & \multicolumn{2}{c|}{20} & \multicolumn{2}{c|}{25} & \multicolumn{2}{c|}{30} & \multicolumn{2}{c}{35} \\
        \hline
        \hline
     & Random & Entropy & Random & Entropy & Random & Entropy & Random & Entropy & Random & Entropy & Random & Entropy \\
    \hline 
    \hline
        Car                   & 31,940 & 32,488 & 42,308 & 42,942 & 56,415 & 53,760 & 71,209 & 64,451 & 88,131 & 74,933 & 108,562 & 82,911 \\ \hline
        Pedestrian            & 20,356 & 24,448 & 30,636 & 31,994 & 40,901 & 39,679 & 46,442 & 48,129 & 54,062 & 58,708 & 61,281 & 62,752 \\ \hline
        Barrier               & 7,915 & 15,224  & 24,166 & 20,335 & 28,904 & 22,117 & 34,338 & 28,117 & 38,903 & 34,791 & 44,906 & 38,639 \\ \hline
        Truck                 & 7,972 & 6,128   & 11,467 & 10,184 & 14,354 & 15,555 & 18,267 & 19,871 & 21,503 & 22,796 & 25,908 & 25,926 \\ \hline
        Traffic Cone          & 3,767 & 6,165   & 10,283 & 8,921  & 12,539 & 10,225 & 15,628 & 13,028 & 18,584 & 15,179 & 20,584 & 18,007 \\ \hline
        Trailer               & 2,562 & 1,635   & 2,779 & 2,977   & 3,801 & 5,750   & 5,580 & 7,658   & 6,448 & 8,237   & 7,486 & 9,591 \\ \hline
        Bus                   & 1,698 & 1,574   & 2,172 & 2,447   & 2,729 & 3,112   & 3,774 & 3,808   & 4,496 & 4,556   & 5,475 & 5,084 \\ \hline
        Construction Vehicle  & 1,262 & 1,401   & 2,138 & 2,253   & 2,877 & 2,903   & 3,678 & 3,634   & 4,595 & 4,366   & 5,145 & 4,752 \\ \hline
        Bicycle               & 762 & 954       & 1,468 & 1,427   & 2,090 & 1,750   & 2,378 & 2,042   & 2,659 & 2,508   & 2,967 & 2,917 \\ \hline
        Motorcycle            & 1,539 & 802     & 1,016 & 1,364   & 1,400 & 1,749   & 1,852 & 2,255   & 2,489 & 2,721   & 2,875 & 3,095 \\ \hline
    \end{tabular}
    \end{adjustbox}
    \label{tab:merged_samples}
\end{table*}

\begin{figure*}
    \centering
\includegraphics[width=0.9\linewidth]{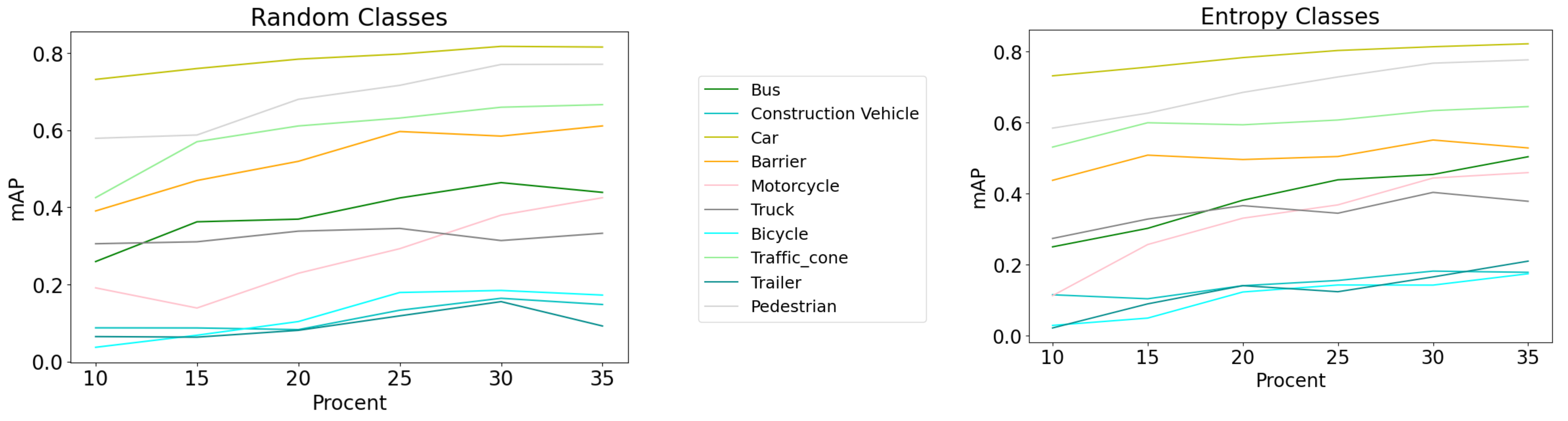}
    \caption{Overview of per-class results, with Random Sampling on left and Entropy Querying on right. While the ordering of classes remains intact and nearly identical to the frequency of appearance of respective classes in the dataset, under entropy sampling, the margin between best and worst performing classes decreases.}
    \label{fig:ClassCompare}
\end{figure*}

\begin{figure*}
    \centering
    \includegraphics[width=.40\textwidth]{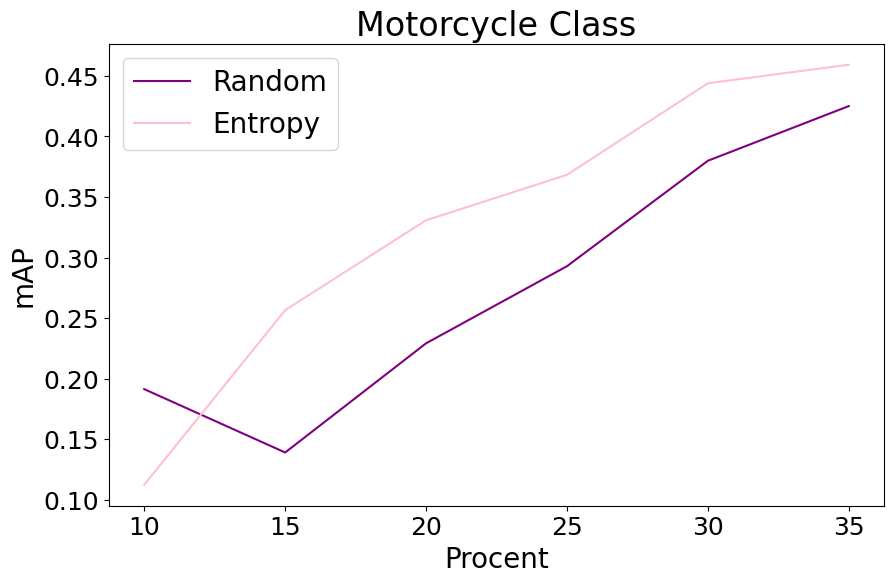}
    \includegraphics[width=.40\textwidth]{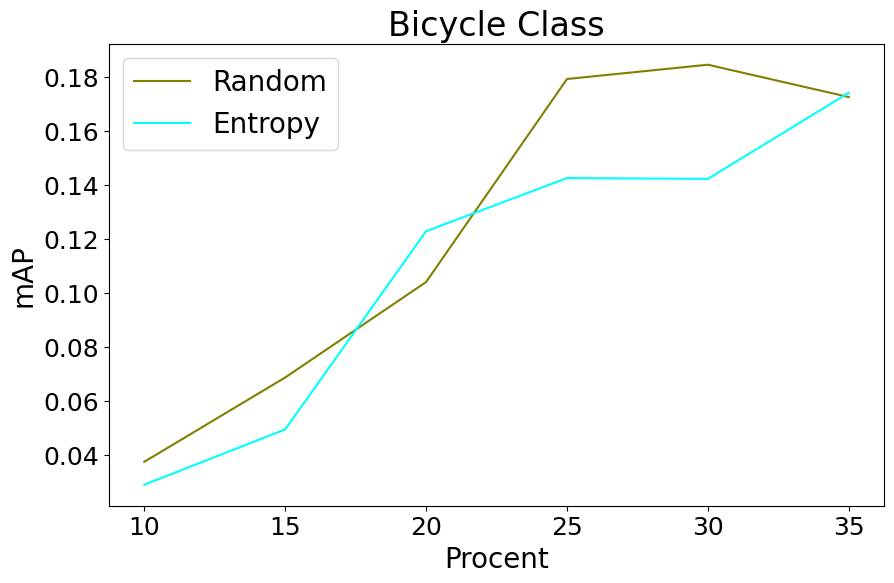}
    \includegraphics[width=.40\textwidth]{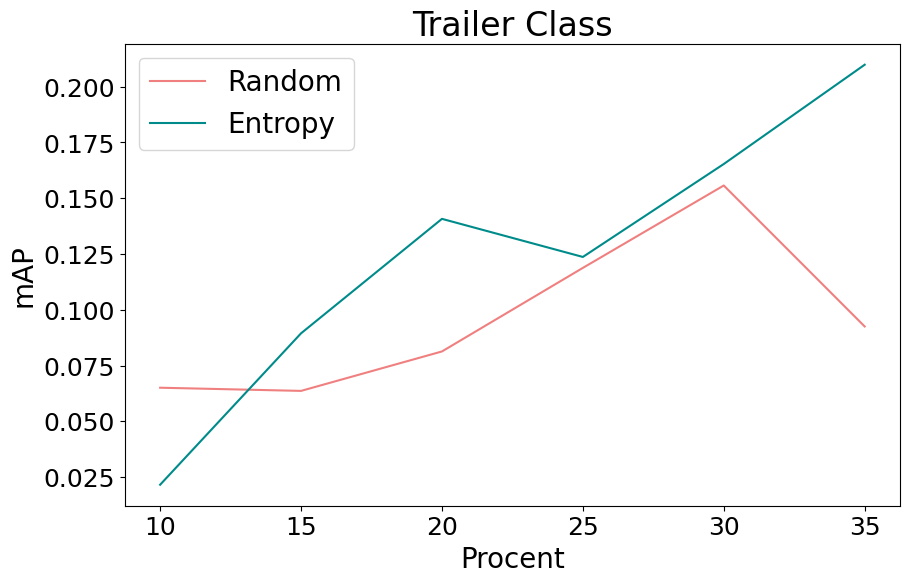}
    \includegraphics[width=.40\textwidth]{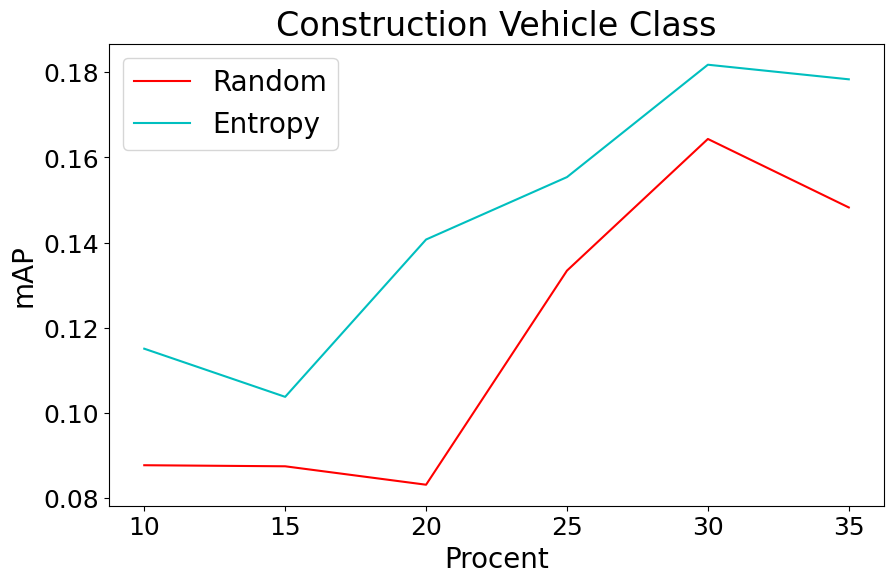}
    \includegraphics[width=.40\textwidth]{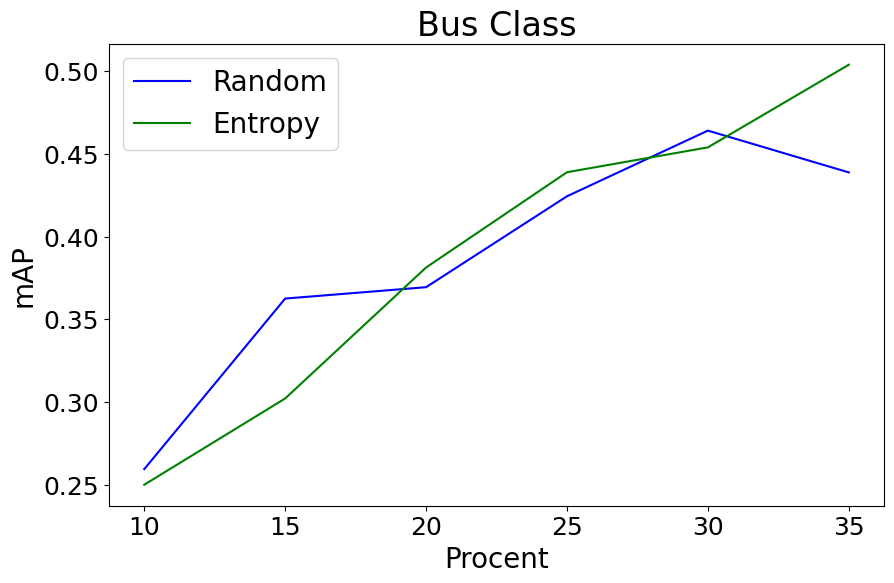}
    \includegraphics[width=.40\textwidth]{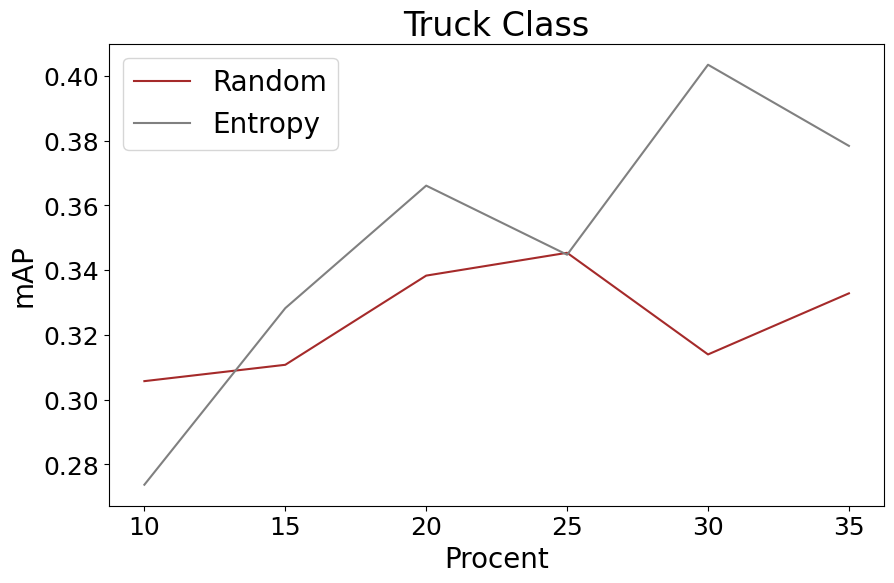}
    \caption{Class-separated analysis of mAP performance between random sampling and entropy query active learning. Entropy query active learning shows a tendency to outperform random sampling on mAP, shown on the six minority classes in these graphs.}
    \label{fig:Classgraphs1}
\end{figure*}

Table \ref{tab:freq} describes the class frequencies of appearance in the nuScenes dataset. We collapse the pedestrian class to contain adults, children, construction workers, those using personal mobility devices, wheelchairs, or strollers, and wearing construction or police uniforms. From Figure \ref{fig:ClassCompare}, we observe that the ordering of classes by highest-to-lowest mAP approximately matches the ordering of class appearance in Table \ref{tab:freq} (car, pedestrian, traffic cone, barrier, truck, bus, motorcycle, construction vehicle, trailer, bicycle). While this ordering is preserved by active learning, we notice that the gap between the lowest mAP and greatest mAP is smaller under active learning, and progressively tightens as more data is added to the pool. Class-specific comparisons are illustrated in Figure \ref{fig:Classgraphs1}. In general, entropy-driven active learning shows improved precision over random selection on all classes, especially beyond early data pool sizes. The margin of performance varies by class.

We make a few observations over these class performances. Most of the worst performing classes (trailer, construction vehicle, bicycle, motorcycle) perform better under entropy sampling than in random sampling. The trailer class performed the worst in random sampling and a little better in entropy sampling, and when looking at Table \ref{tab:merged_samples}, it can be observed that entropy sampling focuses on querying trailer data for every round. The Construction Vehicle class is another class which did not do well in either entropy or random sampling, however, we again see in the table that entropy sampling still outperforms random sampling by a small margin in all rounds, even though the random sampling method draws more examples of this class beyond 30\%, suggesting that the active learning algorithm was not finding better ``informative" samples beyond this point (corroborated by random sampling's greater sampling amount still not besting the performance of entropy querying). As a more classic case, in the motorcycle class, for the initial round the mAP result for this class is comparable to the lowest accuracies observed in other classes. But, under entropy querying, there is a rapid growth in the amount of samples present for this class and as a result the mAP performance consistently increases as the training pool grows. 

\begin{figure*}
    \centering
    \includegraphics[width=.9\textwidth]{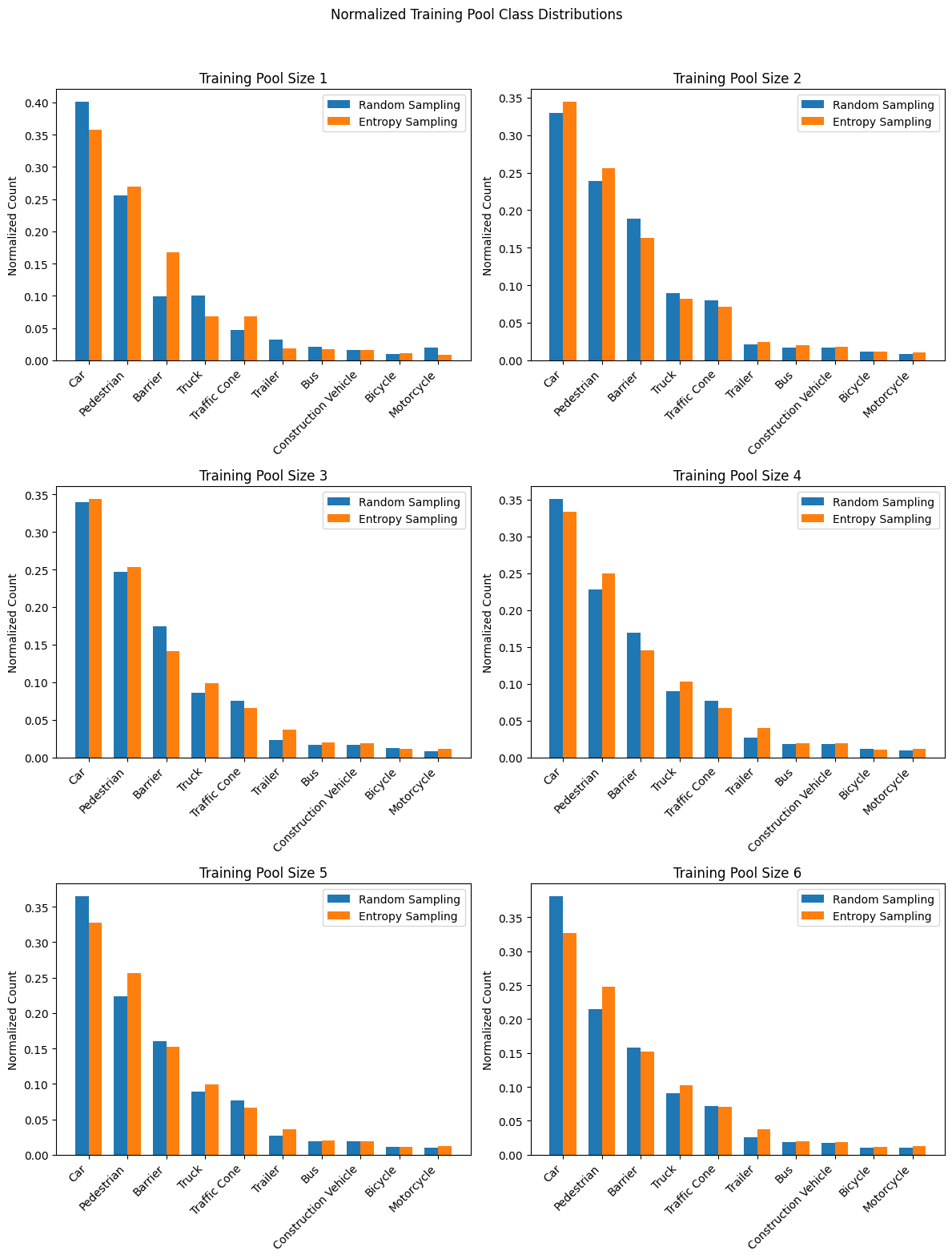}
    \caption{Distribution of samples among classes for each method, varying with training pool size. Entropy sampling methods tend to reduce the addition of majority class (car) samples to the training pool, opting instead to distribute this budget towards the remaining classes. Note that the columns for the respective methods are normalized, as sample sizes will not necessarily sum to the same value since sampling is performed at the scene level, and different scenes may have different numbers of objects.}
    \label{fig:hists}
\end{figure*}

To what extent does entropy querying resolve uncertainty by corrective sampling of minority classes? As shown in Figure \ref{fig:hists}, for each class in each graph, the entropy-driven method tends to pull the distribution to the right toward underrepresented classes as the training pool size increases. We observe the margin between methods for the majority class (car) being widened as the active learning method samples larger pool sizes, with this difference being distributed among the minority classes. The non-normalized data values are presented in Table \ref{tab:merged_samples}.

\section{Concluding Remarks}

Based on the observed results, it is evident that the integration of the entropy querying method with the Birds-Eye-View Fusion model constitutes a favorable combination, demonstrating the effectiveness of active learning. 

One limitation of this analysis is the robustness of results, containing a single method with six iterations of training, each comprising six epochs. To address this limitation, it is recommended that future testing and analysis involve a more extensive approach, where several runs would be conducted for each method. Taking an average of these runs would yield more reliable and comprehensive results. Additionally, the decision to increase the number of epochs from six to ten in future experiments is motivated by the anticipation that a more distinct pattern which more closely matches the fine-tuned state of the art performance of such models may emerge with extended training. Specifically, this adjustment also aligns with the amount of epochs used in the BEVFusion paper, facilitating better direct comparisons with their outcomes.

\subsection{Future Research in Active Learning}

Future research unexplored in active learning in this field includes the learning of query policies directly from autonomous driving tasks and data, instead of relying on handcrafted policies. This could be done using deep reinforcement learning approaches to learn the query policy in the active learning framework. Because the query selection has been shown as a decision process, reinforcement learning can be applied to learn the query \cite{fang2017learning}. Query strategies learned by reinforcement learning have been shown to outperform the heuristic selection methods such as uncertainty sampling and random sampling in a natural language processing task \cite{fang2017learning}. However, to the best of our knowledge, such data-driven query strategies have not been explored in autonomous driving. This is especially important considering the necessity of such systems to efficiently adapt to new environments \cite{lew2022safe, vallon2022data}. \\

\noindent In conclusion, the findings in this research give an affirmation that entropy querying effectively samples the most informative instances from classes with lower accuracies and limited available data, showcasing its utility in the active learning framework, encouraging the adoption of active learning approaches to simultaneously reduce annotation costs and increase data efficiency in learned models for autonomous driving tasks.

\section*{Acknowledgments}
The authors would like to acknowledge the support of Qualcomm through the Qualcomm Inovation Fellowship, and thank mentors Per Siden and Varun Ravi for their valuable feedback. 

\footnotesize
\bibliographystyle{IEEEtran}
\bibliography{biblio}

\end{document}